# In Quest of Image Semantics:
# Are We Looking for It Under the Right Lamppost?

Emanuel Diamant

VIDIA-mant, P.O. Box 933, 55100 Kiriat Ono, Israel

emanl@012.net.il

**Abstract.** In the last years we witness a dramatic growth of research focused on semantic image understanding. Indeed, without understanding image content successful accomplishment of any image-processing task is simply incredible. Up to the recent times, the ultimate need for such understanding has been met by the knowledge that a domain expert or a vision system supervisor have contributed to every image-processing application. The advent of the Internet has drastically changed this situation. Internet sources of visual information are diffused and dispersed over the whole Web, so the duty of information content discovery and evaluation must be relegated now to an image understanding agent (a machine or a computer program) capable to perform image content assessment at a remote image location. Development of Content Based Image Retrieval (CBIR) techniques was a right move in a right direction, launched about ten years ago. Unfortunately, very little progress has been made since then. The reason for this can be seen in a rank of long lasting misconceptions that CBIR designers are continuing to adhere to. I hope, my arguments will help them to change their minds.

## 1   Introduction

There is a widespread assumption that we owe the current uprise of the interest in image semantics to the unprecedented success of the Internet and World Wide Web deployment, which made huge amounts of information (including lots of visual information) easily accessible and affordable over the entire world. Certainly, this assertion is not correct. Understanding image information content was always considered a crucially important prerequisite for any image-processing task. Any successful accomplishment of such a task cannot be even imagined without high-level knowledge mediation, or in other words, without understanding image semantics and specific meaning of its constituting parts.

   The "place" of image understanding in the whole frame of image information processing was set up more than 25 years ago by the classical works of Treisman [1], Marr [2], and Biederman [3]. Treisman's bottom-up/top-down processing paradigm is regarded as the most recognized incarnation of the idea. According to it, processing of image information content is assumed to be an interaction and an interplay of two inversely directed image processing streams. One is an unsupervised, bottom-up directed process of initial image information pieces discovery and localization (the so-called low-level processing stream). The other is a supervised, top-down directed process, which conveys the rules and the knowledge needed to guide the linking and binding of the disjoint low-level information pieces into perceptually meaningful image objects. That is the high-level processing stream associated with image understanding and cognitive image perception.

While the idea of low-level processing from the very beginning was obvious and intuitively appealing (therefore, even today the mainstream of image processing is occupied mainly with low-level pixel-oriented computations), the essence of high-level processing was always obscure, mysterious, and undefined. The paradigm said nothing about the roots of high-level knowledge origination or about the way it has to be incorporated into the introductory low-level processing. Until now, however, the problem was usually bypassed by capitalizing on the expert domain knowledge, adapted to each and every application case. It is not surprising, therefore, that the whole realm of image processing has been (and continues to be) fragmented and segmented according to high-level knowledge competence of the domain experts. For this reason we have today: medical imaging, aerospace imaging, infrared, biologic, underwater, geophysics, remote sensing, microscopy, radar, biomedical, X-ray, and so on "imagings".

The advent of the Internet, with huge volumes of information (including various forms of visual information) scattered over the web, raised an urgent demand for more general means of image semantics recovery, capable to handle visual information in a human-like intelligent manner and at remote image locations. However, deprived of any reasonable sources of the needed high-level knowledge, (and trapped by the tenets of bottom-up/top-down image processing paradigm), computer vision designers are forced to proceed only in one possible direction – to try to derive the high-level knowledge from the available low-level information pieces.

Some theoretical works in biological and computer vision have been done in order to support and to justify this enterprise. Two approaches are prevalent: chaotic attractors approach ([4], [5], [6], [7], [8]), and saliency attention map approach ([9], [10], [11], [12]). Both presume a Shannon-like sense of information, (which is indeed natural for a low-level bottom-up image processing arrangement). Both are computationally expensive. Both definitely violate the basic assumption about the principal role of high-level knowledge in bottom-up low-level processing. But the pressure from Internet providers and users is so high, the lack of other possibilities to reach the needed information is so desperate that no other options are left to CBIR designers. Therefore, all contemporary CBIR developments are continuing to move forward without bothering themselves about the lack of common sense or a firm theoretical justification for their keen and inexhaustible efforts.

## 2  The haste is devil's accommodation

As it was already mentioned above, the urgent need to handle images on the Web in a human-like intelligent manner, while the duties of this management are delegated to an autonomously performing remote agent, has inspired a wave of extensive research and development activities known as CBIR developments. The wave is lasting about a decade long time period. A palpable uprise in the activities has happened since the work got the auspice of the EU IST Programme. A series of excellent follow up reports, depicting the state-of-the-art and the progress made at different program stages, have been regularly published during this time, [13], [14], [15], [16], [17], [18], [19], [20], [21]. There is no need, thus, to repeat and to scrutinize again their accounts. Much more interesting would be to take the bulk of this matter and to use it in order to analyze and to understand the reasons for a suspiciously long delay in achieving the enterprise final goals.

A repeating motif that crowns almost every one of the above mentioned reports is: Despite the efforts and hard work invested in CBIR development, "more work needs to be done in order to be able to automatically extract objective semantics from Web pages, which would allow the retrieval of images on the basis of high-level concepts", ([16], page 61).

It seems that the phrase in the quotation marks already contains the clue for the above raised question. We mean a specific part of the phrase: "to automatically extract objective semantics". We will ignore for a while the doubtful definition "objective semantics" (usually, semantics is considered to be subjective). We will concentrate on the expression "semantics extraction", which explicitly and implicitly present in everyone of the reports just mentioned above, as well as in many other published documents relevant to the subject. This expression reliably represents the general way of thinking dominating today the field of content-based image retrieval research and development. According to it, content-based image retrieval philosophy unconditionally assumes that image semantics can be extracted from a given image (in one way or another), despite all known and well-recognized difficulties, [52].

Logically, what follows from this is: semantics can be extracted from an image only if it belongs to it, that is, if it is an image property, (resident, present, native to an image).

But this is not true! This contradicts all theories (e.g., Hermann von Helmholtz [22]) and even common sense asserts, which claim that images get their semantic meaning as a result of an act of image interpretation. Who carries out this interpretation? The answer can be only one: the human observer who watches the image. Nobody else can do this instead. What follows from this immediately, is that semantics is an exceptionally human's property. It is human's duty and undeniable privilege to provide the image with its semantics.

This changes the situation radically.

Certainly, I am not the first who paid attention to this peculiarity – to the best of my knowledge, it was Santini et al [23] who coined the notion: images are endowed with their semantics. Recently, Hudelot et al [24] have reinvented and reiterated this definition. However, the rest of the research community continues to adhere to the faulty assumption that semantics is an inherent property of an image and thus can be extracted from it or otherwise treated as a built-in property of image data.

Changing the minds and accepting a different point of view will definitely shake up the CBIR design philosophy. From now on, it should be tied and rooted in the neurobiological and neurophysiological studies of human brain, not in pure mathematical and statistical analyses. We will continue to scrutinize the possible implications of this shake up, but before that we would like to examine the rest of the basic assumption underpinning the CBIR designs today.

What we have in mind is the classical formula that the high-level image semantics has to be derived from the available low-level information features. A glance on a list of some representative references will confirm that I do not exaggerate the ubiquity of this attitude in the contemporary CBIR designs, [25], [26], [27], [28]. (Please, remember that these are examples that emphasize only a specific low-level–to–high-level processing passage. The bulk of the evidence is in the regular reports already mentioned above).

A hidden logical inconsistency can be revealed also in this assumption: in their everyday lives people percept the observable objects always as a whole. Semantics is granted to an object always in its entirety, not to its low-level components or features. For a long time, evidence accumulated from psychophysical studies has shown that people can analyze complex visual scenes in extremely short time intervals, lower than 100ms. At such short times only imprecise and coarse-scale representations could be taken into account, any analysis of fine-scale low-level image features is simply impossible, [29], [30].

It will be interesting to recall that an "opposition" to the low-level bottom-up processing (which has always inspired the CBIR designs) has come into being almost simultaneously with the inauguration of the bottom-up processing idea. Navon's "forest before the trees" hypothesis [31] and Chen's "global-to-local topological model" of perceptual organization [32] (published, respectively, in 1977 and 1982) have declared almost 30 years ago that "global topological perception is prior to the perception of other featural properties" [33]. The problem is that all these important insights have happened in biological vision research, while computer vision developers are indifferent to such trinkets.

Now, I have some difficulties to explain why the concept of the Semantic Gap, so widespread and popular among CBIR developers, is the next faulty item in my list of CBIR design misconceptions. In the light of what was already said above, it seems to me that the semantic gap simply does not exist. Definitely, I would be in trouble if I would have to point out where is the semantic gap between the moon of Alabama and the famous Brecht/Weill's song about it. So, I will leave this theme undeveloped, and will shift to the next section of our discussion.

## 3  The time is rape to change your minds

David Marr was the first who has approached vision as an information processing computational process. Although in the late 70s it seemed to him that vision system computations are engaged only with low-level processing. In [34] he writes: "In the early steps of the analysis of an image, the representations used depend more on what it is possible to compute from an image than on what is ultimately desirable".

I appreciate Marr's insights, which have ruled computer vision development almost for 30 years, but I would like (I dare) to propose a different view on the subject. Considering visual information processing, I would like to propose the following definition of the notion "visual image information": Image information is comprised (like all other forms of information) of two parts. The first is the meaningless physical (objective) information, which is a description of visible data structures discernable in an image. The second – is a human produced interpretation of this information. That means, a high-level description of the meaning of the physical information present in an image, which essentially is the image semantics.

I came across the problems of visual information treatment about five years ago. Some results of my investigation are already published, [35], [36]. Some results I am trying to push now. Taking the proposed definition as a start point, I would like to embark on a more reasonable CBIR design approach.

### 3.1 Physical information representation

How biological vision systems acquire and process their input information I have learned quite late, when I had already developed my own approach to the entangled and intriguing problem of image information content discovery and evaluation. I looked for some justifications for my results, and was pleased to find out that such justifications are indeed existent. It was really surprising, because my solution was far away from biology. In my research, I was inspired by the insights of Solomonoff's Inference Theory, Chaitin's Algorithmic Information Theory, and Kolmogorov's Complexity Theory, [35]. Capitalizing on them, I have developed my own way to handle image information content initially represented as raw image data. In simple words my approach looks as follows:

Kolmogorov has combined the theory of computation and a combinatorial approach to information theory in a proposal for an objective and absolute definition of the information contained in an individual finite object. (Contrary to the Shannon's definition, which gives an integrated, averaged over the whole signal ensemble measure of the contained information.) Following Kolmogorov's insights, I have proposed to define the information content of an image as a set of descriptions of the visible image data structures. The Kolmogorov's theory prescribes that such descriptions must be created in a hierarchical and recursive manner. That is, starting with a most generalized and simplified description of image structure, the process has to proceed in a top-down manner to the lower description levels where more and more fine information details can be further elaborated.

A practical algorithm, implementing this idea, has been devised, and its schema is depicted in the Figure 1. As it follows from the schema, to get the introductory generalization, the input image is initially squeezed to a small size of approximately 100 pixels. The rules of this shrinking operation are very simple and fast: four non-overlapping neighbour pixels in an image at level $L$ are averaged and the result is assigned to a pixel in a higher ($L$+1)-level image. Then, at the top of the shrinking pyramid, the image is segmented, and each segmented region is labeled. Since the image size at the top is significantly reduced and since in the course of the bottom-up image squeezing a severe data averaging is attained, the image segmentation/classification procedure does not demand special computational resources. Any well-known segmentation methodology will suffice. We use our own proprietary technique that is based on a low-level (local) information content evaluation and consequent merging of zone borders, [35], but this is not obligatory.

From this point on, the top-down processing path is commenced. At each level, the two previously defined maps (average region intensity map and the associated label map) are expanded to the size of an image at the nearest lower level. Since the regions at different hierarchical levels do not exhibit significant changes in their characteristic intensity, the majority of newly assigned pixels are determined in a sufficiently correct manner. Only pixels at region borders and seeds of newly emerging regions may significantly deviate from the assigned values. Taking the corresponding current-level image as a reference (the left-side unprocessed raw image), these pixels can be easily detected and subjected to a refinement cycle. In such a manner, the process is subsequently repeated at all descending levels until the segmentation/classification of the original input image is successfully accomplished.

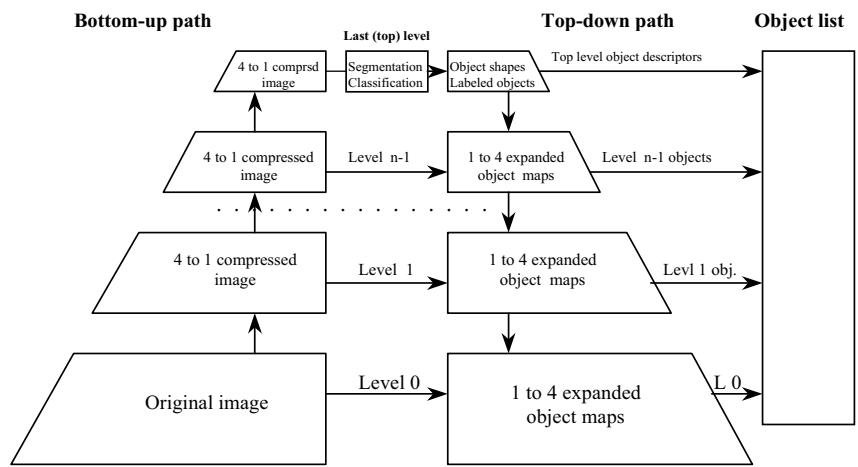

**Fig. 1.** The Schema of the approach

At every processing level, every image sub-region (just recovered or an inherited one) is registered in the objects' appearance list, which is the third constituting part of the proposed schema. The registered object-region parameters are the available simplified object's attributes, such as size, center-of-mass position, average object intensity and hierarchical and topological relationship within and between the objects ("sub-part of…", "at the left of…", etc.). They are sparse, general, and yet specific enough to capture the object's characteristic features in a variety of descriptive forms. This way, a generalized and a simplified description of representative image structures is conserved and carefully recreated at each processing level. Thus, a very effective and optimal (in the sense of the number of recovered sub-region) image segmentation methodology is devised and developed.

What is absolutely clear, is that image information revealed in such a manner refers only to the objective visible image data structures discernable in the image. (That is why I call it "physical information"). It is absolutely independent and dissociated from any high-level semantic knowledge about the image. That opens a unique opportunity to reconsider the way in which we have previously approached image semantics and to try to treat it in a proper separated and independent manner.

**3.2 Semantic knowledge base representation**

As we have already defined earlier, image physical information is interpreted by a human observer in accordance with his previous knowledge about the things in the surrounding world. This knowledge, accumulated in human's head during his entire life span, is usually called human's world ontology. Various aspects of ontology design, use, and amelioration are now a hot topic of the ongoing research in the field of knowledge representation and management. Classical publications related to the subject are well known, [37], [38], [39]. CBIR designers are also deeply involved in attempts to incorporate ontology-based approaches into CBIR architectures, [40], [41], [42], [43], [44], [45], [46], [47], [48]. That is only a partial list, the real outcome is much more extensive. Therefore again, I will not review here the available publications. Since the basic philosophy of the existing approaches and my are so different, I would like to concentrate on the consequences, which follow from such a difference.

First of all, my aim is to resolve the problems of a computer-based remote agent performing autonomously on the Web and expected to interpret (to understand) previously unseen images. For this reason, the agent must be provided with a replica of human ontology, which will facilitate these agent attempts (to endow images on the Web with their semantics). However, being placed in such general and unconstrained conditions, even a human will be unable to fulfill such duties. Even humans are often limited in their knowledge, and perform well only in known and familiar surroundings. Why that should be different for an artificial creature? Our agent, thus, has to be provided (at least at the beginning) with a very restricted set of partial ontologies, which will cover only a limited number of encountered situations. The widespread trend to cover an extended spectrum of image categories looks to me absolutely unrealistic.

Even the restricted knowledge, which is required for agent's functioning, is initially not in its disposal. This knowledge must be appropriately picked up, processed, encoded, arranged and finally in one way or another "shared" (as ontology people call this) between the agent and its supervisors. The agent itself is unable to carry out such tasks. Therefore, it is the designer's duty and the designer's (or system supervisor, or agent's copartner) responsibility to make all these tasks to be accomplished. That, certainly, contradicts the general trend of other ontology systems designs, which presume that systems must be apt to learn and to adapt themselves independently to the changing environment (e.g., new images belonging to different knowledge domains). Obviously, I deny the need for such qualities. I think that the widespread requirement for machine learning capabilities is strongly exaggerated. After all, humans are acquiring their knowledge not in a machine learning fashion, but in a declarative and instructive manner. The knowledge is transferred from a parent to a child, from a teacher to a student, from an instructor to a trainee only in a declarative supervised manner. Recently, it has been discovered that that is not an exceptionally human trait, animals also transfer knowledge between them in a teacher-pupil fashion [49]. So, why it should be different in our case? As a good teacher, which provides his students with deliberately prepared reference material, agent's supervisor, appropriating the agent for a new task, must provide it with a new suitable ontology.

In this regard, a question of a suitable learning language or, more specifically, a question of a language that would best fit the purposes of ontology creation comes into consideration. Today, ontology designers prevalently adhere to the so-called Declarative Language, a sort of a formal language suitable for machine information exchange and machine reasoning. I am not sure that that is good choice for us. Humans in their practice of intercommunication and knowledge exchange (including various levels of education and learning) manage pretty well with their natural language. Of course, this language is not so well formalized, it is fuzzy and imprecise, but usually we don't feel any inconvenience with this. The natural language extremely well serves our needs in communications, knowledge sharing and exchange, learning, reasoning, and problem solving. Anthropomorphous goals of our agent (that has to endow images with their semantics) assume that its knowledge base (supporting and enabling such activity) should also be a human like machinery. That is, the ontology descriptions to be used must be implemented in a natural language.

After all, semantics is an encoded and memorized experience gained in course of human interaction with the surrounding world. Semantics understanding is a recall to this memory in attempt to find a suitable similar situation to capitalize on it in further reasoning and action planning. If we agree that the memorized descriptions could be implemented in words of a natural language, than ontological descriptions that we are speaking about can be seen as a set of narrative chronics (to take account for a time dimension, usually imprudently lost, when we are dealing with images). The equivalent of the human's Long Term Memory, where the ontology descriptions are to be memorized, can be seen then as a large file system, known to us and familiar from memory organization in contemporary computers. May be, an analogue with Object-oriented Programming Environment will be even more appropriate. In any case, this way of ontology description creation and handling looks more natural than the currently used approaches.

### 3.3 How an image can get its semantics

As we just defined above, semantics is an encoded and memorized description of human's experience gained as a result of human interaction with the surrounding world. It is not known for sure, but it is quite possible that human's ontology descriptions are realized as narrative compositions of natural language words. In the previous sub-section we have drawn a parallel between a linguistic description and an ontological description, commonly used today in knowledge representation and management. I would like to continue and to extend this analogy. Linguistic labels for image objects can be seen as equivalents to concept classes, and object sub-parts as equivalents to concept sub-classes. In the regular ontology descriptions, classes as well as subclasses, are usually augmented with attributes, which depict some physical properties of the conceptualized entities represented by a given class (or a sub-class).

Now we must recall how a raw image is initially preprocessed in our proposed approach, (see sub-section 3.1). The physical image information is represented finally as a hierarchical description list of visually distinguishable image sub-regions. One can easily see a striking similarity between the hierarchical description of image sub-regions and the hierarchy of class/sub-class attributes in a scene-related ontology. It is now the designer's duty and responsibility to establish the similarity measures and rules linking between equivalent image objects and their labels. Obviously, all this must be done manually, as a part of the explicit declarative knowledge transfer (learning) mode, announced as the principal modus operandi of our system.

In such a way, an agent can now label the image objects. That is, initially endow them with their semantics. This is only the first step in an image semantics recovery procedure. Having image objects labeled, further, more complex semantic descriptions can be derived now, climbing on the linguistic ontology ladder, and thus pawing a way to a more advanced textual image annotation.

Performing this partial goal, it must be remembered that topological interrelations between image parts (objects) and their subparts (sub-objects), information about coarse/fine parts configurations is critically important for the accomplishment of image semantics endowment, [29], [33], [51]. In this regard, it must be especially emphasized that the proposed physical information representation scheme (sub-section 3.1) provides unprecedented opportunities for exploiting such topological attributes in course of image objects identification (recognition) and subsequent labeling. All these topological relations can be easily depicted and manually linked to the agent's scene ontology.

## 4  Some concluding remarks

In this paper, I have presented an uncommon image understanding and image semantics recovery approach, which today is a vexed topic in many image-handling related research and development enterprises.

I posit that the traditional bottom-up/top-down approach does not hold more. That physical image information can be extracted from an image in an unsupervised top-down manner independent from any high-level knowledge about image content. What follows from this is that high-level concepts are not involved in low-level image processing, and any semantic gap between them does not exist. Semantics is not a property of an image, it is a property of a human observer that watches the image. The observer proceeds image semantics in accordance with his knowledge about the things in the outer world. This knowledge is usually called the ontology of the world. A visual robot or an agent can operate like a human, endowing images with semantics. To enable such activity, it must be furnished with a replica of a human ontology. The latter has not be entirely full (as in humans), but it must be modular, concise, and specific enough about user's view on the particular task in hand. Such a replica can not be achieved in a usual machine-learning fashion, but it must be step-by-step developed and delivered to the agent's disposal by the agent's human supervisor. Such peculiarities sometimes blur the comprehension of the things proposed in this approach. I hope I was lucky this time to make my point clear enough.

# References

1. A. Treisman and G. Gelade, "A feature-integration theory of attention", *Cognitive Psychology*, vol. 12, pp. 97-136, Jan. 1980.

2. D. Marr, "Vision: A Computational Investigation into the Human Representation and Processing of Visual Information", Freeman, San Francisco, 1982.

3. I. Biederman, "Recognition-by-Components: A Theory of Human Image Understanding", Psychological Review, vol. 94, no. 2, pp. 115-147, 1987.

4. M. S. Bartlett and T. J. Sejnowski, "Learning viewpoint invariant face representations from visual experience in an attractor network", *Network: Computation in Neural Systems*, vol. 9, no. 3, pp. 399-417, 1998.

5. Y. Amit, M. Mascaro, "Attractor Networks for Shape Recognition", *Neural Computation*, vol. 13, no. 6, pp. 1415-1442, June 2001.

6. P. E. Latham, S. Deneve, A. Pouget, "Optimal computation with attractor networks", *Journal of Physiology – Paris*, vol. 97, pp. 683-694, 2003.

7. K. McRae, "Semantic Memory: Some insights from Feature-based Connectionist Attractor Networks", Ed. B. H. Ross, The Psychology of Learning and Motivation, vol. 45, 2004. Available: http://amdrae.ssc.uwo.ca/.

8. C. Johansson and A. Lansner, "Attractor Memory with Self-organizing Input", *Workshop on Biologically Inspired Approaches to Advanced Information Technology (BioADIT 2005)*, LNCS, vol. 3853, pp. 265-280, Springer-Verlag, 2006.

9. L. Itti, C. Koch, and E. Niebur, ''A model of saliency-based visual attention for rapid scene analysis", *IEEE Transactions on Pattern Analysis and Machine Intelligence*, vol. 20, No. 11, pp. 1254 – 1259, 1998.

10. A. Shokoufandeh, I. Marsic, S. J. Dickinson, "View-based object recognition using saliency maps", *Image and Vision Computing*, vol. 17, pp. 445-460, 1999.

11. S. Treue, "Visual attention: the where, what, how and why of saliency", *Current Opinion in Neurobiology*, vol. 13, pp. 428-432, 2003.

12. L. Itti, "Models of Bottom-Up Attention and Saliency", In: *Neurobiology of Attention*, (L. Itti, G. Rees, J. Tsotsos, Eds.), pp. 576-582, San Diego, CA: Elsevier, 2005.

13. V.N. Gudivada, V.V. Raghavan, "Content-Based Image Retrieval Systems", *Computer*, vol. 28, iss. 9, pp. 18-22, September 1995.

14. C. Colombo, A. Del Bimbo, and P. Pala, "Semantics in Visual Information Retrieval", IEEE Multimedia, vol. 6, iss. 3, pp. 38-53, July-September 1999.

15. A.W.M. Smeulders, M. Worring, S. Santini, A. Gupta, and R. Jain, "Content-Based Image Retrieval at the End of the Early Years", *IEEE Transactions on Pattern Analysis and Machine Intelligence*, vol. 22, No. 12, pp. 1349-1380, December 2000.

16. M.L. Khefri, D. Ziou, A. Bernardi, "Image Retrieval From the World Wide Web: Issues, Techniques, and Systems", *ACM Computing Surveys*, vol. 36, no. 1, March 2004, pp. 35-67.

17. J. Calic, N. Campbell, S. Dasiopoulou and Y. Kompatsiaris, "An overview of multimodal video representation for semantic analysis", *2nd European Workshop on the Integration of Knowledge, Semantic and Digital Media Technologies (EWIMT-2005)*, 29 November – 1 December, 2005, London. Available: www.cost292.org/pubs/ewimt05.php.

18. Muscle Project, "WP5: State of the Art Report: Image and Video Processing for Multimedia Understanding", Edited by N. Boujemaa, H. Houissa, H. Bischof, Sept. 2004. Available: http://www-rocq.inria.fr/imedia/Muscle/WP5.

19. Muscle Project, "WP5 scientific activity report", October 2005. Available from Muscle WP5 web page: http://www-rocq.inria.fr/imedia/Muscle/WP5.

20. aceMedia Project, "aceMedia Annual Report", November 2005, Available from aceMedia web page: http://www.acemedia.org/aceMedia/files/.

21. M.S. Lew, N. Sebe, C. Djeraba, R. Jain, "Content-based Multimedia Information Retrieval: State of the Art and Challenges", In *ACM Transactions on Multimedia Computing, Communications, and Applications*, February 2006.

22. "Visual perception", From Wikipedia, http://en.wikipedia.org/wiki/Visual_perception.

23. S. Santini, A. Gupta and R. Jain, "Emerging Semantics Through Interaction in Image Databases", *IEEE Transactions of Knowledge and Data Engineering*, vol. 13, No. 3, pp. 337-351, May-June 2001.

24. C. Hudelot, N. Maillot, and M. Thonnat, "Symbol Grounding for Semantic Interpretation: From Image Data to Semantics", Proceedings of the IEEE International Workshop on Semantic Knowledge in Computer Vision, Beijing, China, Oct. 16, 2005, (in association with ICCV 2005). Available: http://www-sop.inria.fr/orion/personnel/Nicolas.Maillot/.

25. Xiang Sean Zhou, T.S. Huang, "CBIR: From low-Level Features to High-Level Semantics", *Proceedings SPIE*, vol. 3974, pp. 426-431, San Jose, CA, January 24-28, 2000. Available: http://www.ifp.uiuc.edu/~xzhou2/.

26. A. Mojsilovic and B. Rogowitz, "Capturing image semantics with low-level descriptors", In Proc. Int. Conf. Image Processing (ICIP-01), Thessaloniki, Greece, October 2001, pp. 18-21.

27. C. Zhang and T. Chen, "From Low Level Features to High Level Semantics", In: *Handbook of Video Databases: Design and Applications*, by Furht, Borko/ Marques, Oge, Publisher: CRC Press, October 2003.

28. D. Depalov, T. Pappas, D. Li, and B. Gandhi, "Perceptually Based Techniques for Semantic Image Classification and Retrieval", *Proceedings SPIE*, vol. 6057, pp. 354-363, San Jose, CA, January 2006.

29. P. G. Schyns and A. Oliva, "From blobs to boundary edges: Evidence for time and spatial scale dependent scene recognition", *Psychological Science*, v. 5, pp. 195 – 200, 1994. Available: http://cvcl.mit.edu/publications.htm/.

30. A. Oliva and A. Torralba, "Building the Gist of a Scene: The Role of Global Image Features in Recognition", In: *Progress in Brain Research, vol. 155: Visual Perception*, 2006. Available: http://cvcl.mit.edu/publications.htm/.

31. D. Navon, "Forest Before Trees: The Precedence of Global Features in Visual Perception", *Cognitive Psychology*, 9, pp. 353-383, 1977.

32. L. Chen, "Topological structure in visual perception", *Science*, 218, pp. 699-700, 1982.

33. L. Chen, "The topological approach to perceptual organization", *Visual Cognition*, vol. 12, no. 4, pp. 553-637, 2005.

34. D. Marr, "Representing visual information: A computational approach", *Lectures on Mathematics in the Life Science*, vol. 10, pp. 61-80, 1978.

35. E. Diamant, "Searching for image information content, its discovery, extraction, and representation", *Journal of Electronic Imaging*, vol. 14, issue 1, January-March 2005. Available: http://www.vidiamant.info.

36. E. Diamant, "Does a plane imitate a bird? Does computer vision have to follow biological paradigms?", In: De Gregorio, M., et al, (Eds.), *Brain, Vision, and Artificial Intelligence,* First International Symposium Proceedings. LNCS, vol. 3704, Springer-Verlag, pp. 108-115, 2005. Available: http://www.vidiamant.info.

37. N. Guarino, "Understanding, Building, And Using Ontologies", *International Journal of Human and Computer Studies,* vol. 46, pp. 293-310, 1997. Available: http://www.loa-cnr.it/Papers/.

38. T.R. Gruber, "Toward Principles for the Design of Ontologies Used for Knowledge Sharing", In: *Formal Ontology in Conceptual Analysis and Knowledge Representation*, Kluwer Academic Publishers, Available: http://kls-web.stanford.edu/authorindex/Gruber.

39. M. Uschold and M. Gruninger, "ONTOLOGIES: Principles, Methods and Applications", *Knowledge Engineering Review*, vol. 11, No. 2, pp. 93-155, 1996.

40. A. Benitez, Shih-Fu Chang, "Automatic Multimedia Knowledge Discovery, Summarization and Evaluation", Avail.: http://www.ee.columbia.edu/dvmm/publications/.

41. P. Bouquet, F. Giunchiglia, F. van Harmelen, L. Serafini, and H. Stuckenschmidt, "C-OWL: Contextualizing Ontologies", *Second International Semantic Web Conference (ISWC-2003)*, LNCS vol. 2870, pp. 164-179, Springer Verlag, 2003.

42. I. Kompatsiaris, Y. Avrithis, P. Hobson and M. Strintzis, "Integrating knowledge, semantics and content for user-centered intelligent media services: the aceMedia project", *Proceedings of Workshop on Image Analysis for Multimedia Interactive Services (WIAMIS'04)*, Portugal, 2004. Available: http://www.image.ntua.gr/~iavr/.

43. Y. Kompatsiaris, et al, "Semantic Annotations of Images and Videos for Multimedia Analysis", *Second European Semantic Web Conference ESWC 2005,* LNCS vol. 3532, pp. 592-607, Springer Verlag, 2005. Available: http://www.iti.gr/db.php/publications/.

44. M. Wallace, Y. Avrithis, G. Stamou and S. Kollias, "Knowledge-Based Multimedia Content Indexing and Retrieval", In: *Multimedia Content and the Semantic Web*, Edited by G. Stamou and S. Kollias, John Wiley & Sons Publishers, 2005, pp. 299-338.

45. D. Vallet, M. Fernandez, and P. Castells, "An Ontology-Based Information Retrieval Model", *2nd European Semantic Web Conference (ESWC 2005)*, LNCS vol. 3532, pp. 455-470, Springer Verlag, 2005.

46. N. Simou, et al, "An Ontology Infrastructure for Multimedia Reasoning", *International Workshop VLBL 05*, Sardinia, Sept. 2005. Available: http://www.image.ntua.gr/~iavt.

47. N. Maillot, M. Thonnat, and A. Boucher, "Towards ontology based cognitive vision", *Machine Vision and Applications (MVA)*, vol. 16, No. 1, pp. 33-40, December 2004.

48. A. Maedche, "Emergent Semantics for Ontologies", *IEEE Intelligent Systems*, vol. 17, issue 1, pp. 85-86, Jan/Feb 2002.

49. N. Franks, T. Richardson, "Teaching in tandem-running ants", *Nature*, 439, p. 153, 12 January 2006.

50. A. Jaimes, "Human Factors in Automatic Image Retrieval System Design and Evaluation", *Proceedings of SPIE, Internet Imaging VII*, vol. 6061, pp. 25-33, San Jose, CA, January 2006.

51. L. Hollink, G. Nguyen, G. Schreiber, J. Wielemaker, B. Wielinga, and M. Worring, "Adding Spatial Semantics to Image Annotations", *4th International Workshop on Knowledge Markup and Semantic Annotation at ISWC'04*, 2004. Available: http://www.cs.vu.nl/~laurah/.

52. M. Naphade and T. S. Huang, "Extracting Semantics From Audiovisual Content: The Final Frontier in Multimedia Retrieval", *IEEE Transactions on Neural Networks*, vol. 13, No. 4, pp. 793-810, July 2002.